\documentclass[runningheads]{llncs}

\usepackage[year=2026,ID=6]{eccv}

\usepackage{eccvabbrv}

\usepackage{graphicx}
\usepackage{booktabs}
\usepackage[accsupp]{axessibility}  
\usepackage{multirow}

\usepackage{amsmath,amsfonts,bm}

\def\eqref#1{equation~\ref{#1}}

\def\1{\bm{1}}

\DeclareMathAlphabet{\mathsfit}{\encodingdefault}{\sfdefault}{m}{sl}
\SetMathAlphabet{\mathsfit}{bold}{\encodingdefault}{\sfdefault}{bx}{n}

\usepackage{hyperref}
\usepackage{orcidlink}

\begin{document}

\title{Mise-en-Scène: Implicit Layout Emergence \\in Diffusion Transformers for Human-AI Design Co-Creation} 

\titlerunning{Mise-en-Sc\`ene}

\author{
Zipeng Xu \and
Ryan Murdock \and
Umberto Michieli
}

\authorrunning{Z.\ Xu et al.}

\institute{
Canva Research\\
\email{\{zipeng,ryanmurdock,umbe\}@canva.com}
}

\maketitle

\begin{abstract}

Automating graphic design synthesis from user-provided elements requires both a coherent overall composition and the exact preservation of each asset. 
Existing methods predict a layout as explicit bounding-box coordinates with a language model and then paste the assets into it, which separates spatial planning from visual synthesis and tends to produce rigid, mis-scaled compositions. We instead ask whether the layout can emerge implicitly inside a pretrained image-editing diffusion transformer. We present \textit{Mise-en-Scène}, a two-stage framework. In the first stage, a diffusion transformer adapted with a small, knockout-selected LoRA drafts a complete design in which the arrangement of the elements emerges jointly with the rendered canvas. 
In the second stage, a deterministic match-and-place step moves the original high-resolution assets to the drafted positions, which guarantees exact asset fidelity and yields an editable, layered design that a designer can keep refining rather than a flat image.
Notably, a minimal adaptation of the pretrained transformer already suffices, without the specialized conditioning machinery commonly introduced for multi-element generation. On the large-scale PrismLayersPlus benchmark, the designs produced by \textit{Mise-en-Scène} are the closest to the ground truth in perceived quality among all compared methods, by a wide margin over both an LLM layout planner and a specialized layout transformer, while our match-and-place stage bridges the remaining fidelity gap to the ground-truth composites.

\end{abstract}

\section{Introduction}

\begin{figure}[ht]
  \centering
  \includegraphics[width=.95\linewidth]{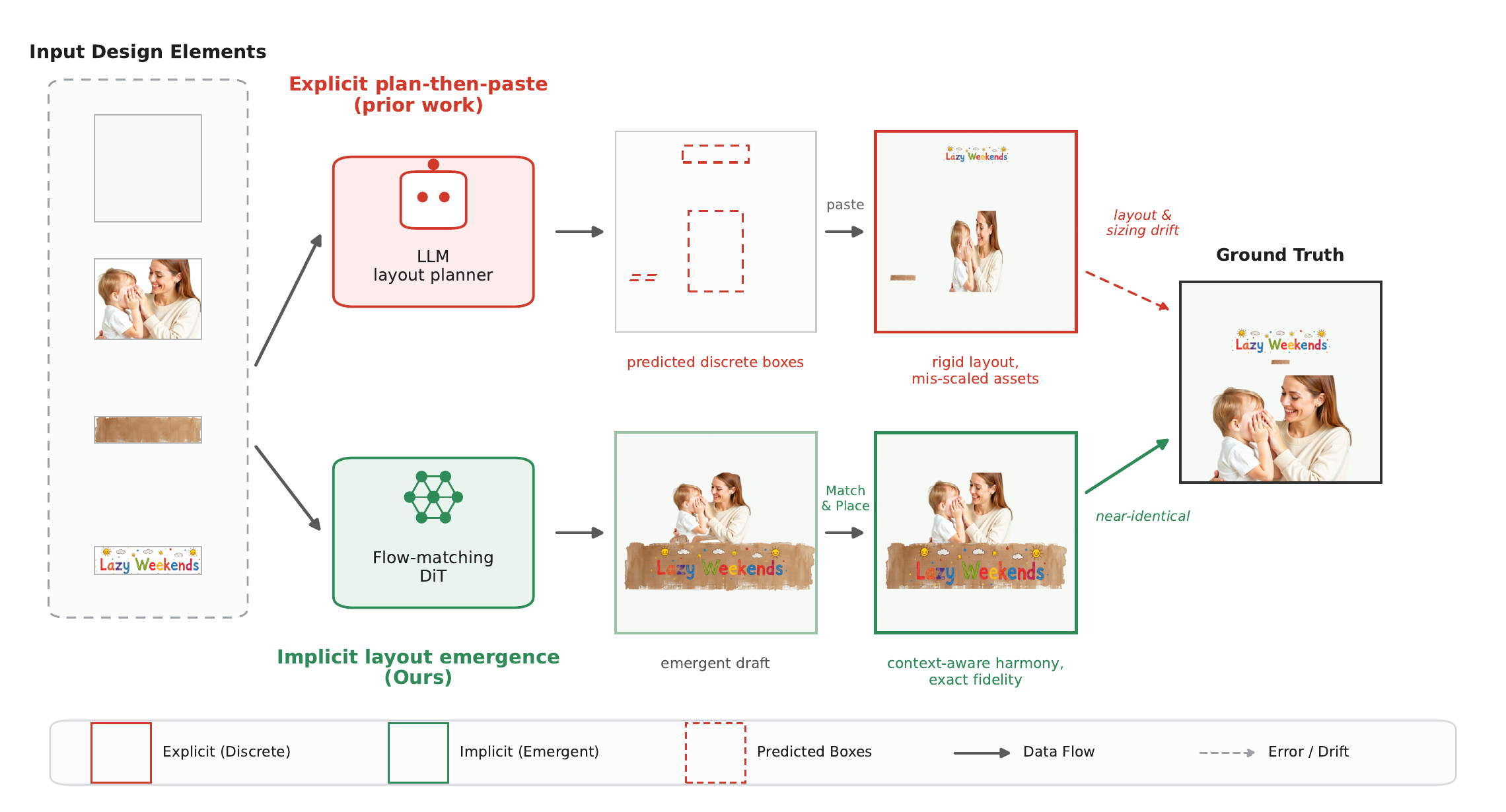}
  \caption{\textbf{Two paradigms for element-conditioned design.} Given the same design elements provided without spatial context (left), the explicit plan-then-paste pipeline (top) has an LLM predict discrete bounding boxes and pastes the assets into them, which often mis-scales elements and yields rigid layouts. Our \textit{Mise-en-Sc\`ene} (bottom) lets the layout emerge implicitly inside a flow-matching diffusion transformer and then restores exact pixels with a deterministic match-and-place step, producing a composition close to the ground truth. The boxes in the top row are the planner's real predictions, and every image is a real output on the same sample.}
  \label{fig:teaser}
\end{figure}
Visual design is fundamentally a collaborative process between creative intent and aesthetic execution
~\cite{yamaguchi2021canvasvae,
jia2023cole,
inoue2024opencole,
pu2025art,
peng2025bizgen,
chen2025prismlayersopendatahighquality,
zhang2025creatiposter,
chen2025postercraft,
shabani2024composer,
cheng2025graphic,
goyal2025design,
zhang2025creatidesign}.
In practical scenarios, designers rarely start from a blank canvas; rather, they are driven by a predefined set of unstructured visual elements—such as product cutouts, brand assets (e.g., logos), and specific textual slogans—which must be composed into a coherent and visually compelling design. 
Transforming these isolated assets into an aesthetically balanced layout demands a profound understanding of spatial hierarchy and visual harmony.
Traditionally, this task has been carried out manually by professional designers using tools such as Canva or Photoshop.
Providing designers with automatically generated candidates that they can adopt or refine has the potential to significantly improve creative efficiency and accessibility, particularly for non-expert users.
This motivates the task of design image generation conditioned on given visual elements, which aims to synthesize complete designs guided by structured visual inputs~\cite{shabani2024composer,
cheng2025graphic,
goyal2025design,
zhang2025creatidesign}.

Existing approaches to element-based design generation predominantly rely on decoupled, multi-stage pipelines. A common paradigm, exemplified by recent state-of-the-art frameworks, leverages Large Language Models (LLMs) to explicitly predict spatial bounding boxes as discrete coordinates, followed by image synthesis or post-hoc compositing~\cite{lin2023layoutprompter, tang2024layoutnuwa, mahajan2026gist}. However, this LLM "plan-then-paste" approach suffers from fundamental limitations. LLMs, inherently biased toward 1D sequential text reasoning, often lack a native understanding of dense 2D spatial relationships and fine-grained visual aesthetics~\cite{rahmanzadehgervi2025visionlanguagemodelsblind}. Forcing a continuous visual composition task into discrete coordinate tokens introduces quantization errors and creates a severe modality gap. In addition, the planner never observes the renderer, so it cannot account for how the elements will actually be drawn, and placement errors propagate to the final design, leading to suboptimal visual harmony.
Figure~\ref{fig:teaser} contrasts this explicit plan-then-paste pipeline with the implicit alternative we pursue.

In this work, 
we propose a paradigm shift by posing a fundamental scientific question: \textit{Can we bypass explicit LLM-based layout prediction and directly unlock the implicit layout expertise and aesthetic composition capabilities natively embedded within large-scale pre-trained diffusion models?} 
Generative foundation models, through training on billions of image-text pairs, have already internalized rich continuous priors regarding structural balance, text-background contrast, and contextual object placement.
To validate this, we introduce \textit{Mise-en-Sc\`ene}, which turns a pretrained image-editing diffusion transformer into an element-conditioned design generator using only a small, knockout-selected LoRA. The elements enter the joint attention stream of the model as ordinary visual tokens, and their spatial arrangement emerges jointly with the rendered canvas, with no explicit coordinate predictor.
In our experiments, the extra conditioning components introduced by prior multi-element work bring no benefit, and the simplest adaptation works best; our final model is therefore deliberately minimal.

While the DiT backbone excels at establishing a globally coherent layout and aesthetic harmony, stochastic diffusion sampling inherently struggles to guarantee the pixel-perfect reconstruction required for strict brand asset fidelity. To address this, our framework adopts a compositional post-processing strategy. We append a deterministic, non-learned Match-and-Place refinement module to the generation pipeline.
A vision-language model grounds each input element in the generated draft, returning the box where the model placed it, and we paste the original asset layer at that box. 
This reads out the layout that the diffusion model produced and re-renders it from the original pixels, so the final design preserves every asset exactly and remains an editable, layered document rather than a flat image.
Because the output stays editable, the generated design is a starting point the designer can refine rather than a fixed result, supporting a human-AI co-creation workflow.

Our contributions can be summarized as follows:
\begin{itemize}
    \item We introduce a generative paradigm for element-conditioned design synthesis that shifts from explicit LLM-based layout planning to implicit layout emergence within a pretrained diffusion transformer that we adapt with only a small LoRA.
    \item We introduce a deterministic Match-and-Place scheme that reinstates the original assets as separate layers, ensuring $100\%$ visual identity preservation and returning a fully editable layered design rather than a flat image.
    \item Extensive experiments on the large-scale PrismLayersPlus~\cite{chen2025prismlayersopendatahighquality} benchmark show that our designs are the closest to the ground truth in perceived quality among all compared methods, substantially ahead of both an LLM layout planner and a specialized layout transformer.

\end{itemize}

\section{Related Work}

\subsection{Graphic Design Synthesis and Composition}
Automating graphic design has evolved rapidly, moving from early latent modeling of vector graphic documents~\cite{yamaguchi2021canvasvae} to multi-stage, multi-modal pipelines. A dominant line of work decomposes a design into hierarchical layers and renders text, foreground, and background sequentially~\cite{jia2023cole, inoue2024opencole}, a paradigm subsequently specialized for infographics~\cite{peng2025bizgen} and high-quality, editable poster synthesis~\cite{zhang2025creatiposter, chen2025postercraft}. The maturation of large-scale transparent multi-layer datasets~\cite{chen2025prismlayersopendatahighquality} and dedicated layered generators~\cite{pu2025art} has further fueled this direction. A parallel thread leverages large multimodal models to evaluate, critique, and iteratively refine designs~\cite{goyal2025design, cheng2025graphic}, and to unify multi-conditional generation within a single diffusion transformer~\cite{shabani2024composer, zhang2025creatidesign}.

Most relevant to us is \emph{element-conditioned} composition, where a fixed set of user-provided assets must be assembled into a finished design. 
LaDeCo~\cite{lin2024elements}, the current state of the art for element-conditioned design composition, casts this as a layered, sequential rendering problem driven by an LLM layout planner, while GIST~\cite{mahajan2026gist} operates as a post-hoc compositor that harmonizes already-placed elements. We argue these methods share a common structural limitation: they enforce a decoupled \emph{plan-then-paste} (or plan-then-composite) pipeline that severs spatial planning from visual synthesis. Because the planner never observes the generative manifold of the renderer, errors compound across stages and global aesthetic harmony is sacrificed for local placement. In contrast, \textit{Mise-en-Scène} treats element-conditioned composition as a single-pass generative process in which layout emerges \emph{implicitly} and jointly with visual harmonization, and invokes explicit coordinates only at the very end to guarantee pixel fidelity.

\subsection{Generative Layout Modeling}
Predicting where elements belong is a long-standing sub-problem of design synthesis. Early approaches modeled bounding-box distributions with GANs~\cite{li2019layoutgan, li2020attributegan, zhou2022CGLGAN, hsu2023posterlayout}, after which diffusion models were adapted to the discrete-continuous nature of layout~\cite{chai2023layoutdm, inoue2023layoutdm, levi2023dlt, hui2023unifying, chen2024towards}. More recently, the field has reframed layout as a sequence-modeling task solved by LLMs and LMMs~\cite{lin2023layoutprompter, tang2024layoutnuwa, seol2024posterllama, yang2024posterllava, chen-etal-2024-textlap, horita2024retrievalaugmented}, with some works ingesting full multimodal markup documents~\cite{kikuchi2024multimodal}. While such models excel at sequential reasoning, we contend they are structurally mismatched to dense 2D layout: serializing continuous spatial relations into 1D discrete coordinate tokens imposes an unnatural quantization and opens a modality gap between a text-biased planner and a pixel-level synthesizer. A complementary family of grounded generators—\eg GLIGEN~\cite{li2023gligen}—instead injects bounding boxes as explicit conditioning, but still presupposes that coordinates are given rather than discovered. \textit{Mise-en-Scène} departs from both paradigms: rather than predicting or consuming discrete boxes, it resolves spatial configurations natively within the continuous latent space of a diffusion transformer, and recovers explicit coordinates only \emph{a posteriori} for evaluation and fidelity-preserving recomposition.

\subsection{Image-Conditioned Generation and Visual In-Context Learning}
Injecting specific visual elements into diffusion models has been explored extensively. Adapter- and attention-based methods such as IP-Composer~\cite{dorfman2025ip} and part-based concepting frameworks~\cite{richardson2025piece} fuse image embeddings through decoupled cross-attention. With the shift to Diffusion Transformers~\cite{peebles2023dit}, a powerful line of \emph{visual in-context} approaches now treats reference assets as additional tokens within the joint attention stream, unlocking controllability directly from visual prompts via token concatenation, flow matching, and lightweight fine-tuning~\cite{wu2025uno, li2025visualcloze, lhhuang2024iclora, labs2025flux1kontextflowmatching}. 
Of particular relevance, OminiControl~\cite{tan2024ominicontrol, tan2025ominicontrol2} and UNO~\cite{wu2025uno} report that \emph{shifting} the Rotary Position Embedding (RoPE) of conditioning tokens to non-overlapping index ranges helps disambiguate multiple references and stabilize training. We build on the same design-oriented backbone (Qwen-Image-Edit~\cite{wu2025qwenimage}) and evaluate such an offset in our multi-element setting, but find it unnecessary in our experiments: the pretrained backbone already separates the concatenated elements without an explicit positional offset.

A second challenge in multi-element conditioning is \emph{concept bleeding}—the architectural tendency of attention to leak features between subjects~\cite{dahary2024boundedattention}. UNet-era solutions localize cross-attention either by training-time supervision against segmentation masks~\cite{xiao2024fastcomposer} or layout boxes~\cite{wang2024msdiffusion}, or by training-free attention manipulation at inference~\cite{chen2024trainingfreelayout, dahary2024boundedattention}. 
We revisit this idea in the DiT joint-attention regime and experiment with analogous attention-grounded supervision, namely a box-alignment term that steers each element toward its region and a separation term that discourages overlap between elements' attention maps. In our setting this supervision leaves quality essentially unchanged: the joint-attention backbone does not show the concept bleeding that motivated these losses in the UNet era, so we omit them in our final model.

To ensure perfect identity preservation of image elements, our framework recovers an explicit layer composition as a post-process. This connects to the rich literature on layer decomposition and transparent-layer generation, from latent-transparency diffusion~\cite{zhang2024layerdiffuse} and region transformers~\cite{pu2025art} to recent raster-to-layer decomposers~\cite{crello_suzuki2025layerddecomposingrastergraphic, yang2025layerdecomp, yin2025qwenimagelayered}. Rather than learning a decomposition network, we exploit the zero-shot visual grounding ability of modern VLMs~\cite{bai2025qwen25vl} to localize each generated element and deterministically substitute the original high-resolution asset, combining the diffusion model's emergent aesthetics with lossless identity preservation.

\begin{figure}[t]
    \centering
    \includegraphics[width=.985\linewidth]{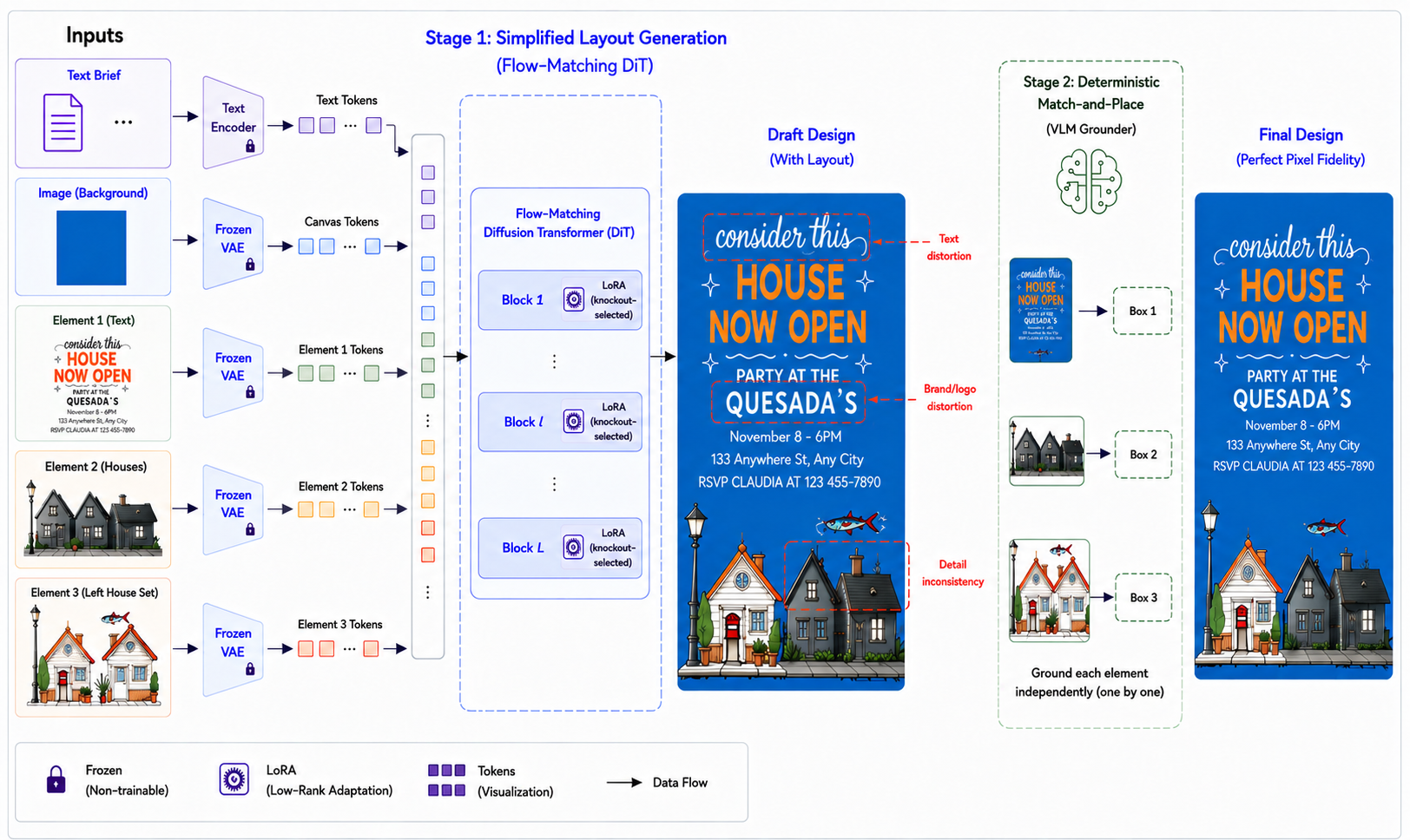}
    \caption{\textbf{Overview of \textit{Mise-en-Sc\`ene}.} The method runs in two stages. In Stage~1 (simplified layout generation), a text brief, a background canvas, and the visual elements are mapped by frozen encoders into a joint token sequence and processed by a flow-matching Diffusion Transformer adapted only through a knockout-selected LoRA, which produces a layout-aware draft design. Because the draft comes from stochastic sampling, it still contains rendering artifacts such as distorted text and warped logos. In Stage~2 (deterministic match-and-place), a VLM grounder locates each original element in the draft independently, and the original high-resolution layers are alpha-composited at the grounded boxes to yield the final design, which keeps the emergent layout while restoring exact, pixel-faithful, editable assets.}
    \label{fig:framework}
\end{figure}

\section{Problem Formulation}
\label{sec:problem_formulation}

We formulate element-conditioned design generation as a unified spatial arrangement and visual compositing task. A training sample is formalized as an ordered tuple:
\begin{equation}
    (c_0, c_1, \dots, c_N, y, \tau, \{b_k\}_{k=1}^N),
\end{equation}
where $c_0$ represents the base background canvas, $c_1, \dots, c_N$ are the transparent foreground element layers (RGBA images), $y$ is the finalized composite design, $\tau$ is a structured text brief, and $b_k = [x^{\min}, y^{\min}, x^{\max}, y^{\max}]$ denotes the ground-truth bounding box of element $k$ within the coordinate space of $y$.

Crucially, the input visual elements $c_{0:N}$ are provided \textit{out of spatial context}. The generative model must independently infer their optimal joint arrangement to learn the conditional distribution $p_\theta(y \mid c_{0:N}, \tau)$. 
The ground-truth boxes $\{b_k\}$ are not used during training; they serve only to evaluate the extracted layout at test time.
During inference, the model implicitly resolves spatial relationships, allowing us to subsequently extract precise per-element placements $\hat{b}_k$ to drive the post-processing renderer.

\section{Method}
\label{sec:method}

Existing components-to-design frameworks typically formulate layout generation as a discrete coordinate prediction problem~\cite{lin2024elements}. This decoupled paradigm severs the intrinsic connection between structural planning and pixel-level harmonization. Building upon the problem formulation in Section~\ref{sec:problem_formulation}, our framework, \textit{Mise-en-Sc\`ene}, operates as a unified two-stage pipeline. First, we achieve implicit layout emergence with a flow-matching Diffusion Transformer (DiT) adapted by a knockout-selected LoRA. Second, we deploy a deterministic match-and-place post-processing stage that guarantees exact pixel fidelity and returns an editable, layered design.

As illustrated in Figure~\ref{fig:framework}, the generative stage denoises the target canvas $y$ while attending jointly to the text brief and the VAE-encoded visual elements. To adapt the model without eroding its pretrained rendering prior, we rely on resolution-agnostic conditioning and a task-specific knockout-guided LoRA, which we detail in the following subsections.

\subsection{Implicit Layout Generation via Flow-Matching DiT}
\label{subsec:implicit_generation}

We build on a large-scale pretrained image-editing DiT. Below we describe how the elements enter its joint attention sequence, which projections we adapt, and the training objective.

\textbf{Joint Sequence Modeling and Resolution-Agnostic Conditioning.}
The network denoises the target canvas $y$ while processing all cross-modal information via a fully connected joint self-attention mechanism. Let $c_k \in \mathbb{R}^{H_k \times W_k \times 4}$ denote the $k$-th visual asset (an RGBA image with four channels). To prevent token explosion from high-resolution inputs and handle elements of varying aspect ratios, each element is dynamically resized to a fixed pixel budget $A$ (where $H'_k \times W'_k \approx A$) prior to being processed by the frozen VAE encoder $\mathcal{E}_{\text{vae}}$. This yields a flattened latent token sequence $\mathbf{z}_k \in \mathbb{R}^{L_k \times D}$, where $L_k = (H'_k / v) \times (W'_k / v)$ and $v$ is the patch size. Concurrently, the text brief $\tau$ is encoded by a text encoder $\mathcal{E}_{\text{text}}$ into language tokens $\mathbf{E}_\tau \in \mathbb{R}^{L_\tau \times D}$. The generative process operates on the concatenated multi-modal joint sequence:
\begin{equation}
    \mathbf{Z} = \big[ \mathbf{E}_\tau,\, \mathbf{z}_0,\, \mathbf{z}_1,\, \dots,\, \mathbf{z}_N,\, \mathbf{z}_y \big] \in \mathbb{R}^{L \times D},
\end{equation}
where $\mathbf{z}_0$ represents the base background canvas, $\mathbf{z}_y \in \mathbb{R}^{L_y \times D}$ represents the noisy target latent, and $L = L_\tau + \sum_{k=0}^{N} L_k + L_y$ is the total sequence length. Placeholders in $\tau$ act as semantic anchors, binding the encoded textual intent directly with the corresponding visual token spans.

\textbf{LoRA Adaptation.}
Rather than adapt the full model, we apply LoRA to a small set of target projections chosen by a causal knockout procedure: starting from adapters on all candidate projections, we disable each in turn, measure the resulting change in the task metric on a held-out set, and keep the $K{=}9$ families whose adaptation matters most, leaving the rest of the backbone frozen. Adapting all LoRA-able projections instead performs comparably (Table~\ref{tab:ablation}), so we prefer this smaller set for its lower parameter count and faster training. This choice follows the spirit of parameter-efficient concept customization~\cite{kumari2023customdiffusion}.

\textbf{Flow-Matching Objective.}
The draft is learned with a standard flow-matching objective. We parameterize the vector field $v_\theta$ to follow the optimal-transport path between the noise distribution $x_0 \sim \mathcal{N}(0, I)$ and the data distribution $x_1$ (the clean target latent), sampling intermediate states on the linear path $x_t = (1-t)\,x_0 + t\,x_1$ under a shifted timestep schedule $t \in [0, 1]$. Writing $\mathbf{Z}_{<y} = [\mathbf{E}_\tau,\, \mathbf{z}_0,\, \dots,\, \mathbf{z}_N]$ for the conditioning tokens (the joint sequence $\mathbf{Z}$ without its target block $\mathbf{z}_y = x_t$), the objective is:
\begin{equation}
    \mathcal{L}_{\text{fm}} = \mathbb{E}_{t, x_1, x_0} \Big[ \big\| v_\theta(x_t, t, \mathbf{Z}_{<y}) - (x_1 - x_0) \big\|_2^2 \Big].
\end{equation}
We use no auxiliary layout losses; the placement is learned end to end from the flow-matching signal alone.

\subsection{Deterministic Match-and-Place}
\label{subsec:match_and_place}

The draft $\hat{y}$ settles the layout, but stochastic sampling does not reconstruct high-frequency content such as small typography and vector logos exactly. We therefore separate layout from final rendering: the draft supplies the arrangement, and a deterministic match-and-place step re-renders the design from the original assets.

\textbf{Grounding by visual matching.} We use a vision-language model (VLM) as a visual grounder. For each element $c_k$, the VLM is given the draft $\hat{y}$ together with the original element image and returns the tightest box in $\hat{y}$ whose content matches $c_k$:
\begin{equation}
    \hat{b}_k = \mathrm{VLM}(\hat{y},\, c_k).
\end{equation}
Each element is grounded independently, so the VLM does not plan a layout; it reads out the layout that the DiT has already produced. This is the essential difference from plan-then-paste baselines, whose boxes come from a planner that never observes the rendered design.

\textbf{Placement.} Every element is grounded in this way, including the base canvas $c_0$, which the VLM localizes just like the foreground layers. Starting from a transparent RGBA canvas $\mathbf{C}^{(-1)}$, we composite the elements back-to-front: for $k = 0, 1, \dots, N$ we resize the original element to its grounded box $\hat{b}_k$ and alpha-composite it onto the running canvas:
\begin{equation}
    \mathbf{C}^{(k)}(p) = \alpha_k(p)\, \tilde{c}_k(p) + \big( 1 - \alpha_k(p) \big)\, \mathbf{C}^{(k-1)}(p),
\end{equation}
where $\tilde{c}_k$ is the resized element and $\alpha_k$ its alpha channel, and the final design is $\mathbf{C}^{(N)}$. Because every element is reinstated as its own layer at a grounded position, the output preserves the exact appearance of each asset ($100\%$ identity preservation) and remains a fully editable, re-arrangeable layered design rather than a flat raster.

\section{Experimental}
\subsection{Experimental Setup}
\label{sec:experimental_setup}

\textbf{Dataset.} We evaluate on PrismLayersPlus~\cite{chen2025prismlayersopendatahighquality}, containing approximately 97K commercial designs across 21 styles. Each sample provides a base canvas, $N$ layer assets, the final composite, and layer-specific bounding boxes. The dataset is split into 78,299 / 9,787 / 9,785 for training, validation, and testing. 
Our main evaluation uses a stratified test set of 1{,}000 designs (TEST1000; 50 per style, up to 4 foreground layers), and the qualitative figures draw from the same test pool. 
Prior layout work is often evaluated on Crello~\cite{yamaguchi2021canvasvae}, but a typical Crello design is built from many small vector and text elements, whereas our task composes a few user-provided image elements (photographs, logos, illustrations); we therefore evaluate on PrismLayersPlus.

\textbf{Text brief.} Each design is paired with a structured brief $\tau$ that describes \emph{what} the design contains rather than where its elements go: a background description, one short phrase per foreground element, and an overall style tag, assembled into a fixed template, ``\texttt{Background: [bg]. Elements - Picture 1: [element 1]; \dots; Picture $N$: [element $N$]. Style: [style].}'' The label \texttt{Picture $k$} corresponds to the $k$-th foreground element, so each text span binds to the matching visual tokens in the joint sequence. We generate the per-element phrases by prompting a large vision-language model (Qwen3.5-VL) on each element in isolation; our prompt ablation (Table~\ref{tab:ablation}) replaces these phrases with the dataset's original captions placed in the same template.

\textbf{Implementation Details.} We build on the Qwen-Image-Edit DiT and inject LoRA (rank 512) into the 9 knockout-selected target projections: the query, key, value, and output projections of the image-stream self-attention (\texttt{to\_q}, \texttt{to\_k}, \texttt{to\_v}, \texttt{to\_out}), the query and output projections of the text stream (\texttt{add\_q\_proj}, \texttt{to\_add\_out}), the image and text feed-forward output projections (\texttt{img\_mlp}, \texttt{txt\_mlp}), and the image modulation projection (\texttt{img\_mod}); the knockout drops the remaining families, 
notably the text-stream key/value and modulation projections. Each conditioning element is resized to a pixel budget of $262{,}144$ ($\approx 512^2$) before the frozen VAE encoder, and we keep up to five conditioning images per sample: the base canvas $c_0$ plus up to $N\,{=}\,4$ foreground elements $c_1,\dots,c_N$.
Training uses a learning rate of $5 \times 10^{-5}$ with cosine decay, 500 warmup steps, and an effective batch size of 48 (distributed across 8$\times$H200 GPUs). We use Classifier-Free Guidance with a 10\% condition-dropout probability and zero the timestep conditioning. In inference we sample for 40 steps with a guidance scale of 3.0 and generate at $\approx 1024^2$ to match the ground-truth resolution. For match-and-place, the Qwen3-VL-8B grounder locates each element independently and returns a box in a normalized $[0,1000]$ space; we denormalize it, resize the original RGBA asset to that box by Lanczos resampling, and alpha-composite the elements back-to-front in their layer order. An element that the grounder fails to locate is left out of the composite.


\textbf{Evaluation Protocol \& Metrics.} For fairness, we align all generation resolutions to the target ground truth ($\approx 1024^2$) and evaluate box geometry in a scale-independent space. Following LaDeCo~\cite{lin2024elements}, we report two families of metrics. \textbf{Geometry (Val, Olap, Align)} measures the validity, element overlap, and global alignment of the extracted layout boxes, each read as closeness to the ground-truth statistics. \textbf{Aesthetics (LVM)} measures design quality with a large multimodal judge (Qwen3-VL-8B) across five criteria, reported as the absolute deviation from the ground truth ($|\Delta\text{GT}|$).

\textbf{Baselines.} We compare against two baselines that we reproduce and retrain on PrismLayersPlus using the authors' official open-source code: 
LaDeCo~\cite{lin2024elements}, the strongest prior approach to this task and a recent LLM-based layout planner, and FlexDM~\cite{inoue2023flexible}, a masked multi-modal layout transformer. Both predict a layout and render it by pasting the original assets, so we pass every method through the \textit{same} VLM grounder and rendering pipeline, ensuring all methods operate under the same evaluation ceiling.

\begin{table*}[t!]
  \centering
  \small
  \caption{Comparison of aesthetic quality and layout geometry on the PrismLayersPlus TEST1000 split. Aesthetic dimensions are rated by Qwen3-VL-8B: Design (i), Content (ii), Typography (iii), Graphics (iv), and Innovation (v). $\text{Ove}_{\mathrm{aes}}$ is the overall aesthetic score and $|\Delta \text{GT}|$ its absolute deviation from the ground truth (lower is better). Geometric metrics report layout Validity (Val), Overlap (Olap), and Alignment (Align) on the extracted boxes. Our method is closest to the ground truth in overall aesthetic score and on every criterion, while all methods reach comparable geometric validity.}
  \label{tab:comprehensive_evaluation}
  \begin{tabular}{@{}lcccccccccccc@{}}
    \toprule
    \multirow{2}{*}{Methods} & \multicolumn{7}{c}{Qwen3-VL Aesthetic Scores ($\uparrow$)} & & \multicolumn{3}{c}{Geometric Metrics} \\
    \cmidrule(lr){2-8} \cmidrule(lr){10-12}
    & (i) & (ii) & (iii) & (iv) & (v) & $\text{Ove}_{\mathrm{aes}}$ & $|\Delta \text{GT}|$ ($\downarrow$) & & Val & Olap & Align \\
    \midrule
    FlexDM~\cite{inoue2023flexible} & 6.71 & 7.04 & 6.64 & 6.88 & 6.04 & 6.66 & 0.371 & & 0.996 & 0.170 & 0.0057 \\
    LaDeCo~\cite{lin2024elements} & 6.91 & 7.21 & 6.74 & 7.01 & 6.08 & 6.79 & 0.240 & & 0.999 & 0.106 & 0.0057 \\
    \textbf{Ours} & \textbf{7.06} & \textbf{7.36} & \textbf{6.91} & \textbf{7.25} & \textbf{6.35} & \textbf{6.99} & \textbf{0.046} & & 0.997 & 0.112 & 0.0047 \\
    \midrule
    GT & 7.12 & 7.53 & 6.96 & 7.26 & 6.29 & 7.03 & 0.000 & & 0.999 & 0.091 & 0.0068 \\
    \bottomrule
  \end{tabular}
\end{table*}

\begin{figure}[p]
   \centering
   \includegraphics[width=\linewidth]{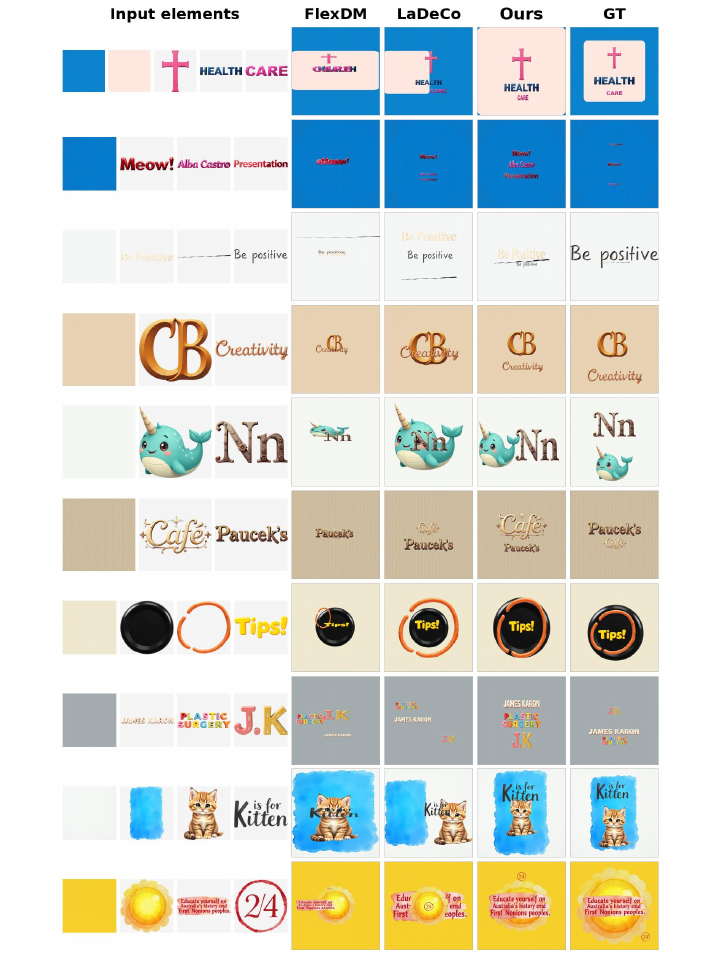}
   \caption{\textbf{Qualitative comparison on the PrismLayersPlus test set.} Each row shows, from left to right, the input design elements provided without spatial context, followed by the composites from FlexDM, LaDeCo, our \textit{Mise-en-Sc\`ene}, and the ground truth. Every method is rendered from its own predicted layout; our column is the match-and-place output. Our designs follow the ground-truth arrangement most closely; see text for a per-row discussion.}
   \label{fig:qualitative}
 \end{figure}

\subsection{Quantitative Results}
\textbf{Aesthetic quality.} Table~\ref{tab:comprehensive_evaluation} shows that \textit{Mise-en-Sc\`ene} produces the most ground-truth-like designs among all methods. 
Its overall aesthetic score (6.99) is the highest of the three methods, ahead of the prior state of the art, LaDeCo (6.79), and of FlexDM (6.66), and the same ordering holds on every individual criterion.
Its deviation from the ground truth is only 0.046, five to eight times smaller than LaDeCo (0.240) and FlexDM (0.371). Resolving the layout inside the continuous diffusion latent space thus yields more design-realistic compositions than pasting assets at coordinates emitted by a language planner.

\textbf{Geometric layout properties.} All methods produce almost entirely valid layouts (Val $\geq 0.996$). Our alignment error (0.0047) is the lowest of the three methods, and our overlap (0.112) is close to LaDeCo (0.106) and to the ground-truth level (0.091), indicating well-formed arrangements rather than degenerate stacking. Our aim is not the most precise box placement but the most design-realistic composition, and the geometric metrics confirm that this realism does not come at the cost of malformed layouts.

\textbf{Cross-judge consistency.} We adopt Qwen3-VL-8B as our primary judge, and additionally report LLaVA-OneVision-7B, the judge used by LaDeCo~\cite{lin2024elements}, which we retain only for comparability as it is a somewhat older model. The ranking by distance to the ground truth is nonetheless stable across both: on LLaVA-OneVision our designs deviate from the ground truth by only 0.003 in overall aesthetic score, versus 0.047 for LaDeCo and 0.093 for FlexDM, mirroring the Qwen3-VL ordering in Table~\ref{tab:comprehensive_evaluation}. Both judges therefore place our results closest to the ground truth by a clear margin, which makes the comparison less dependent on any single evaluator.

\subsection{Qualitative Results}

Figure~\ref{fig:qualitative} compares our designs with FlexDM and LaDeCo on the test split. FlexDM predicts all element positions jointly and tends to overlap and mis-scale assets: it paints the title over the subject in the \emph{Kitten} and \emph{Health Care} rows and shrinks the narwhal to a fraction of its size in the \emph{Nn} row, so its compositions look cluttered and lose text legibility. LaDeCo places the elements more conservatively, but its arrangements are frequently less harmonious than ours: in the \emph{Tips!} row it leaves the orange ring floating beside the button rather than around it, and in the \emph{Nn} row it oversizes the narwhal so that it crowds the letters. Our results integrate the same elements more coherently and follow the ground-truth composition most closely, sizing each asset in proportion and keeping text in uncluttered space, as in the \emph{Tips!}, \emph{Nn}, and \emph{CB Creativity} rows.

Two properties of our pipeline stand out. First, because the layout emerges together with the rendered canvas rather than from pre-committed coordinates, foreground subjects are sized in proportion to the design and titles remain in open space for readability. Second, the match-and-place stage replaces each drafted element with its original asset, so logos and typography are reproduced exactly in the final composite.

\begin{figure}[t!]
  \centering
  \includegraphics[width=.75\linewidth]{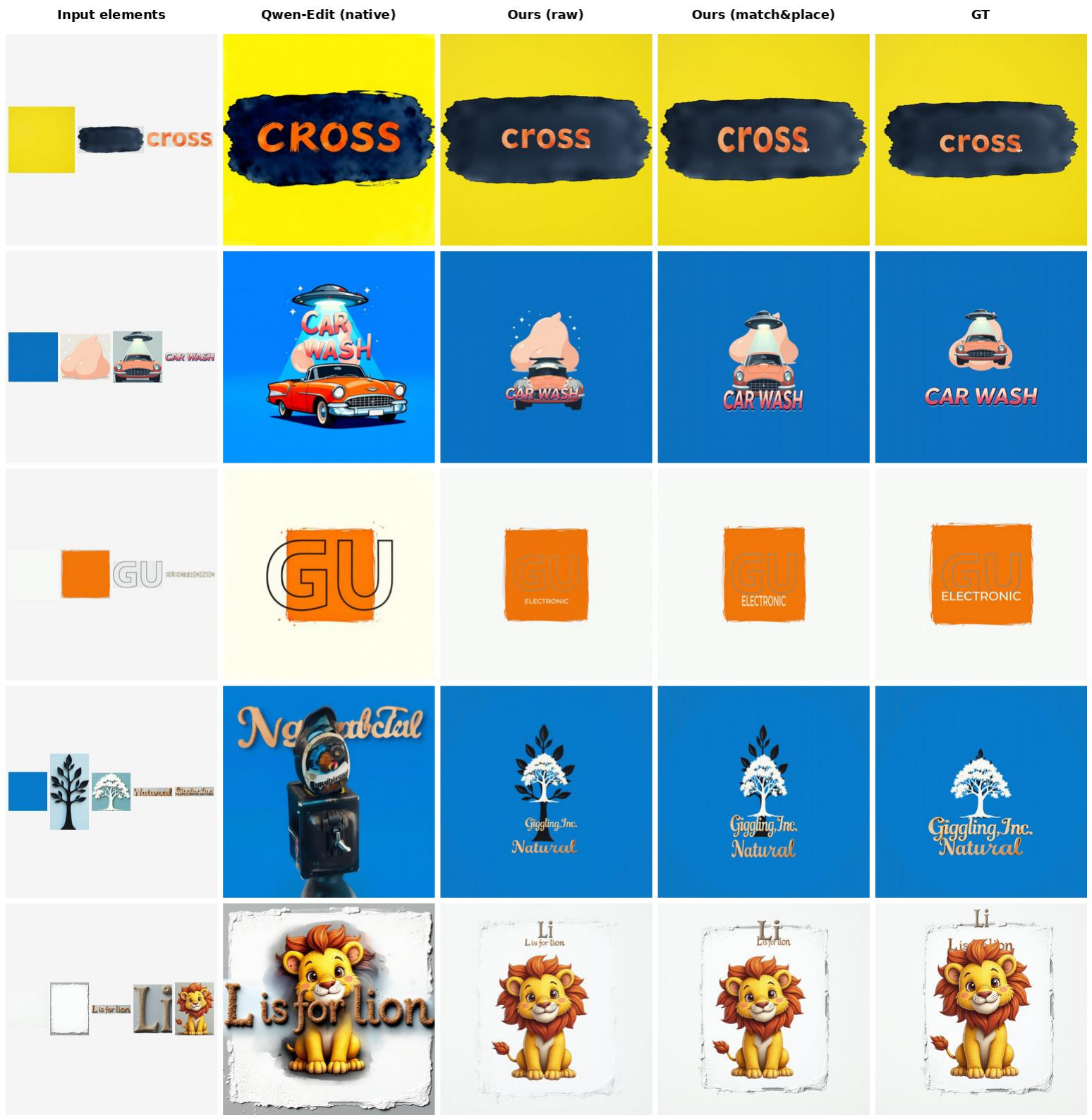}
  \caption{\textbf{Comparison with the zero-shot backbone, and the effect of match-and-place.} For each sample, from left to right: the input elements, the pretrained Qwen-Image-Edit applied zero-shot, our raw diffusion draft, our match-and-place result, and the ground truth. The zero-shot backbone distorts assets with few elements and loses element identity with more, whereas our raw draft fixes the layout and match-and-place restores the original assets for exact fidelity; see text for details.}

  \label{fig:base}
\end{figure}

\subsection{Comparison with the Base Model}
To isolate what \textit{Mise-en-Sc\`ene} contributes, we run the pretrained Qwen-Image-Edit backbone zero-shot on the same task, giving it the same base canvas, element layers, and text prompt as our model and the same inference settings, but without our LoRA. Figure~\ref{fig:base} shows its native output alongside our raw draft, our match-and-place result, and the ground truth. With only a few elements the backbone composes a plausible layout but re-renders the assets in its own style instead of preserving them: in the top row it redraws the ``cross'' wordmark as heavier all-caps type, and for the ``GU'' logo it changes the color while dropping the ``ELECTRONIC'' wordmark. As the number of elements grows it degrades sharply, losing element identity and omitting elements: for the tree-and-text logo it replaces the tree with an unrelated object and renders illegible text, and for the car-wash poster it redraws the car as a different vehicle. Our fine-tuned model composes the same elements reliably, and match-and-place then restores each asset exactly.

\subsection{Ablation Study}

\begin{table*}[t!]
  \centering
  \small
  \caption{\textbf{Ablation study on the PrismLayersPlus TEST1000 split.} We ablate the design choices of our final model: knockout-based selection of the LoRA target modules (vs adapting all target projections, ``Full LoRA''), the VLM prompt rewrite (vs the dataset's raw captions), and the match-and-place stage (vs scoring the raw diffusion draft). Metrics follow Table~\ref{tab:comprehensive_evaluation}.}
  \label{tab:ablation}
  \begin{tabular}{@{}lcccccccccccc@{}}
    \toprule
    \multirow{2}{*}{Model Variants} & \multicolumn{7}{c}{Qwen3-VL Aesthetic Scores ($\uparrow$)} & & \multicolumn{3}{c}{Geometric Metrics} \\
    \cmidrule(lr){2-8} \cmidrule(lr){10-12}
    & (i) & (ii) & (iii) & (iv) & (v) & $\text{Ove}_{\mathrm{aes}}$ & $|\Delta \text{GT}|$ ($\downarrow$) & & Val & Olap & Align \\
    \midrule

    Full LoRA & 7.02 & 7.28 & 6.88 & 7.25 & 6.30 & 6.95 & 0.084 & & 0.999 & 0.116 & 0.0053 \\
    w/o rewrite & 7.02 & 7.29 & 6.89 & 7.22 & 6.28 & 6.94 & 0.090 & & 0.998 & 0.123 & 0.0049 \\
    w/o M\&P & 6.80 & 6.95 & 6.66 & 7.11 & 6.40 & 6.78 & 0.248 & & - & - & - \\
    \midrule
    \textbf{Ours} & \textbf{7.06} & \textbf{7.36} & \textbf{6.91} & \textbf{7.25} & \textbf{6.35} & \textbf{6.99} & \textbf{0.046} & & 0.997 & \textbf{0.112} & \textbf{0.0047} \\
    \bottomrule
  \end{tabular}
\end{table*}

We ablate the design choices of our final model on the PrismLayersPlus TEST1000 split (Table~\ref{tab:ablation}).

\textbf{VLM prompt rewrite.} Replacing the VLM-rewritten prompts with the dataset's raw captions leaves placement and geometry essentially unchanged (Val/Olap/Align within noise), but moves the design further from the ground truth in aesthetic score (Qwen3-VL $|\Delta \text{GT}|$ from 0.046 to 0.090). The rewrite phrases the elements and style in a way that yields a more ground-truth-like design.

\textbf{LoRA target selection.} Adapting all LoRA-able projection families (Full LoRA) performs on par with our 9-family knockout selection, with all differences within noise on the 1000-sample test. We therefore keep the smaller set, which matches full adaptation while training far fewer parameters and at lower cost.

\textbf{Match-and-Place.} Scoring the raw diffusion draft instead of the match-and-place composite lowers aesthetic fidelity sharply, moving $|\Delta \text{GT}|$ from 0.046 to 0.248 on Qwen3-VL, while placement and geometry are unaffected because the boxes are read from the same draft. The \emph{Ours (raw)} and \emph{Ours (match\&place)} columns of Figure~\ref{fig:base} show this step: the drafted elements are replaced by their original pixels, most visibly for small text and logos. 
This shows the division of labor: the diffusion stage sets the layout while match-and-place supplies most of the final aesthetic score, not a cosmetic touch-up.

\subsection{Limitations}
Our approach has some limitations. First, the diffusion draft itself does not yet reach the quality of real designs: match-and-place repairs the high-frequency assets, but the raw generation still trails the ground truth in global styling, and the final result inherits this gap. 
Second, the final placement depends on the VLM grounder; when it mislocalizes an element the error carries into the composite, and coverage is bounded by the grounder's accuracy, which we use off-the-shelf---fine-tuning it on our task would likely raise this ceiling.
Third, the human-AI co-creation setting is realized through the editable, layered output but is not yet validated in an interactive study with designers, and our evaluation focuses on the PrismLayersPlus domain rather than a broad range of design styles. Finally, following common multi-concept and multi-subject generation setups~\cite{kumari2023customdiffusion, wang2024msdiffusion}, each design is conditioned on at most five elements (a base canvas and up to four foreground layers); designs with substantially more elements are outside our current scope. We view these as natural directions for future work.


\section{Conclusion}

We presented \textit{Mise-en-Sc\`ene}, a two-stage framework for element-conditioned design that replaces explicit layout planning with implicit layout emergence inside a pretrained diffusion transformer. A small knockout-selected LoRA is enough to elicit this behavior, and a deterministic match-and-place stage then restores the original assets at the emergent positions, yielding an exact-fidelity, editable design. On PrismLayersPlus, our designs are the closest to the ground truth in perceived quality among all compared methods, favoring design realism over exact coordinate prediction. 
We further find that a minimal adaptation is enough, without the extra conditioning mechanisms usually added for multi-element generation. Because its output is an editable, layered design rather than a fixed image, the framework keeps the designer in the loop, supplying the elements and refining the result, which fits a human-AI co-creation workflow. We hope that implicit layout emergence provides a simple and strong basis for future design co-creation tools.

\bibliographystyle{splncs04}
\bibliography{main}


\end{document}